# New Techniques for Context Modeling

Eric Sven Ristad and Robert G. Thomas
Department of Computer Science
Princeton University
{ristad,rgt}@cs.princeton.edu

## Abstract

We introduce three new techniques for statistical language models: extension modeling, nonmonotonic contexts, and the divergence heuristic. Together these techniques result in language models that have few states, even fewer parameters, and low message entropies.

## 1 Introduction

Current approaches to automatic speech and handwriting transcription demand a strong language model with a small number of states and an even smaller number of parameters. If the model entropy is high, then transcription results are abysmal. If there are too many states, then transcription becomes computationally infeasible. And if there are too many parameters, then "overfitting" occurs and predictive performance degrades.

In this paper we introduce three new techniques for statistical language models: extension modeling, nonmonotonic contexts, and the divergence heuristic. Together these techniques result in language models that have few states, even fewer parameters, and low message entropies. For example, our techniques achieve a message entropy of 1.97 bits/char on the Brown corpus using only 89,325 parameters. By modestly increasing the number of model parameters in a principled manner, our techniques are able to further reduce the message entropy of the Brown Corpus to 1.91 bits/char.[1] In contrast, the character 4-gram model requires 250 times as many parameters in order to achieve a message entropy of only 2.47 bits/char. Given the logarithmic nature of codelengths, a savings of 0.5 bits/char is quite significant. The fact that our model performs significantly better using vastly fewer parameters argues that it is a much better probability model of natural language text.

Our first two techniques – *nonmonotonic contexts* and *extension modeling* – are generalizations of the traditional context model (Cleary and Witten 1984; Rissanen 1983,1986). Our third technique – *the divergence heuristic* – is an incremental model selection criterion based directly on Rissanen's (1978) minimum description length (MDL) principle. The MDL principle states that the best model is the simplest model that provides a compact description of the observed data.

In the traditional context model, every prefix and every suffix of a context is also a context. Three consequences follow from this property. The first consequence is that the context dictionary is unnecessarily large because most of these contexts are redundant. The second consequence is to attenuate the benefits of context blending, because most contexts are equivalent to their maximal proper suffixes. The third consequence is that the length of the longest candidate context can increase by at most one symbol at each time step, which impairs the model's ability to model complex sources. In a nonmonotonic model, this constraint is relaxed to allow compact dictionaries, discontinuous backoff, and arbitrary context switching.

The traditional context model maps every history to a unique context. All symbols are predicted using that context, and those predictions are estimated using the same set of histories. In contrast, an extension model maps every history to a *set* of contexts, one for each symbol in the alphabet. Each symbol is predicted in its own context, and the model's current predictions need not be estimated using the same set of histories. This is a form of parameter tying that increases the accuracy of the model's predictions while reducing the number of free parameters in the model.

As a result of these two generalizations, nonmonotonic extension models can outperform their equivalent context models using significantly fewer parameters. For example, an order 3 n-gram (ie., the 4-gram) requires more than 51 times as many con-texts and 787 times as many parameters as the order 3 nonmonotonic extension model, yet already performs worse on the Brown corpus by 0.08 bits/char.

---

[1] The only change to our model selection procedure is to replace the incremental cost formula $\Delta L_\phi(w, \Sigma', \sigma)$ with a constant cost of 2 bits/extension. This small change reduces the test message entropy from 1.97 to 1.91 bits/char but it also quadruples the number of model parameters and triples the total codelength.

Our third contribution is the divergence heuristic, which adds a more specific context to the model only when it reduces the codelength of the past data more than it increases the codelength of the model. In contrast, the traditional selection heuristic adds a more specific context to the model only if it's entropy is less than the entropy of the more general context (Rissanen 1983,1986). The traditional minimum entropy heuristic is a special case of the more effective and more powerful divergence heuristic. The divergence heuristic allows our models to generalize from the training corpus to the testing corpus, even for nonstationary sources such as the Brown corpus.

The remainder of our article is organized into three sections. In section 2, we formally define the class of extension models and present a heuristic model selection algorithm for that model class based on the divergence criterion. Next, in section 3, we demonstrate the efficacy of our techniques on the Brown Corpus, an eclectic collection of English prose containing approximately one million words of text. Section 4 discusses possible improvements to the model class.

## 2 Extension Model Class

This section consists of four parts. In 2.1, we formally define the class of extension models and prove that they satisfy the axioms of probability. In 2.2, we show to estimate the parameters of an extension model using Moffat's (1990) "method C." In 2.3, we provide codelength formulas for our model class, based on efficient enumerative codes. These codelength formulas will be used to match the complexity of the model to the complexity of the data. In 2.4, we present a heuristic model selection algorithm that adds parameters to an extension model only when they reduce the codelength of the data more than they increase the codelength of the model.

### 2.1 Model Class Definition

Formally, an *extension model* $\phi = \langle \Sigma, D, E, \lambda \rangle$ consists of a finite alphabet $\Sigma$, $|\Sigma| = m$, a dictionary $D$ of contexts, $D \subset \Sigma^*$, a set of available context extensions $E$, $E \subseteq D \times \Sigma$, and a probability function $\lambda : E \to [0, 1]$. For every context $w$ in $D$, $E(w)$ is the set of symbols available in the context $w$ and $\lambda(\sigma|w)$ is the conditional probability of the symbol $\sigma$ in the context $w$. Note that $\sum_{\sigma \in \Sigma} \lambda(\sigma|w) \leq 1$ for all contexts $w$ in the dictionary $D$.

The probability $\tilde{p}(h|\phi)$ of a string $h$ given the model $\phi$, $h \in \Sigma^n$, is calculated as a chain of conditional probabilities (1)

$$\tilde{p}(h|\phi) \doteq \tilde{p}(h_n|h_1 \ldots h_{n-1}, \phi)\tilde{p}(h_1 \ldots h_{n-1}|\phi) \quad (1)$$

while the conditional probability $\tilde{p}(\sigma|h, \phi)$ of a single symbol $\sigma$ after the history $h$ is defined as (2).

$$\tilde{p}(\sigma|h, \phi) \doteq \begin{cases} \lambda(\sigma|h) & \text{if } \langle h, \sigma \rangle \in E \\ \delta(h)\tilde{p}(\sigma|h_2 h_3 \ldots h_n, \phi) & \text{otherwise} \end{cases} \quad (2)$$

The expansion factor $\delta(h)$ ensures that $\tilde{p}(\cdot|h, \phi)$ is a probability function if $\tilde{p}(\cdot|h_2 \ldots h_n, \phi)$ is a probability function.

$$\delta(h) \doteq \frac{1 - \lambda(E(h)|h)}{1 - \tilde{p}(E(h)|h_2 \ldots h_n, \phi)} \quad (3)$$

Note that $E(h)$ represents a set of symbols, and so by a slight abuse of notation $\lambda(E(h)|h)$ denotes $\sum_{\sigma \in E(h)} \lambda(\sigma|h)$, ie., the sum of $\lambda(\sigma|h)$ over all $\sigma$ in $E(h)$.

**Example 1.** Let $\Sigma = \{0, 1\}$, $D = \{\epsilon, \texttt{"0"}\}$, $E(\epsilon) = \{0, 1\}$, $E(\texttt{"0"}) = \{0\}$. Suppose $\lambda(0|\epsilon) = \frac{1}{2}$, $\lambda(1|\epsilon) = \frac{1}{2}$, and $\lambda(0|\texttt{"0"}) = \frac{3}{4}$. Then $\delta(\texttt{"0"}) = \frac{1}{4}/\frac{1}{2} = \frac{1}{2}$ and $\tilde{p}(1|\texttt{"0"}, \phi) = \delta(\texttt{"0"}) \lambda(1|\epsilon) = \frac{1}{4}$.

The fundamental difference between a context model and an extension model lies in the inputs to the context selection rule, not its outputs. The traditional context model includes a selection rule $s : \Sigma^* \to D$ whose only input is the history. In contrast, an extension model includes a selection rule $s : \Sigma^* \times \Sigma \to D$ whose inputs include the past history and the symbol to be predicted. This distinction is preserved even if we generalize the selection rule to select a set of candidate contexts. Under such a generalization, the context model would map every history to a set of candidate contexts, ie., $s : \Sigma^* \to 2^D$, while an extension model would map every history and symbol to a set of candidate contexts, ie., $s : \Sigma^* \times \Sigma \to 2^D$.

Our extension selection rule $s : \Sigma^* \times \Sigma \to D$ is defined implicitly by the set $E$ of extensions currently in the model. The recursion in (2) says that each symbol should be predicted in its longest candidate context, while the expansion factor $\delta(h)$ says that longer contexts in the model should be trusted more than shorter contexts when combining the predictions from different contexts.

An extension model $\phi$ is *valid* iff it satisfies the following constraints:

a. $\epsilon \in D \wedge E(\epsilon) = \Sigma$
b. $\forall w \in D \ [\sum_{\sigma \in E(w)} \lambda(\sigma|w) \leq 1]$
c. $\forall w \in D \ [E(w) = \Sigma \Rightarrow \sum_{\sigma \in E(w)} \lambda(\sigma|w) = 1]$
(4)

These constraints suffice to ensure that the model $\phi$ defines a probability function. Constraint (4a) states that every symbol has the empty string as a context. This guarantees that every symbol will always have at least one context in every history and that the recursion in (2) will terminate. Constraint (4b) states that the sum of the probabilities of the extensions $E(w)$ available in in a given context $w$ cannot sum to more than unity. The third constraint (4c) states that the sum of the probabilities of the extensions $E(w)$ must sum exactly to unity when every symbol is available in that context (ie., when $E(w) = \Sigma$).

**Lemma 2.1** $\forall y \in \Sigma^* \quad \forall \sigma \in \Sigma$
$[\ \tilde{p}(\Sigma|y) = 1 \Rightarrow \tilde{p}(\Sigma|\sigma y) = 1\ ]$

**Proof.** By the definition of $\delta(\sigma y)$.

**Theorem 1** *If an extension model $\phi$ is valid, then* $\forall n \sum_{s \in \Sigma^n} \tilde{p}(s|\phi) = 1$.

**Proof.** By induction on $n$. For the base case, $n = 1$ and the statement is true by the definition of validity (constraints 4a and 4c). The induction step is true by lemma 2.1 and definition (1). □

## 2.2 Parameter Estimation

Let us now estimate the conditional probabilities $\lambda(\cdot|\cdot)$ required for an extension model. Traditionally, these conditional probabilities are estimated using string frequencies obtained from a training corpus. Let $c(\sigma|w)$ be the number of times that the symbol $\sigma$ followed the string $w$ in the training corpus, and let $c(w)$ be the sum $\sum_{\sigma \in \Sigma} c(\sigma|w)$ of all its conditional frequencies.

Following Moffat (1990), we first partition the conditional event space $\Sigma$ in a given context $w$ into two subevents: the symbols $q(w)$ that have previously occurred in context $w$ and those that $\bar{q}(w)$ that have not. Formally, $q(w) \doteq \{\sigma : c(\sigma|w) > 0\}$ and $\bar{q}(w) \doteq \Sigma - q(w)$. We estimate $\lambda_C(q(w)|w)$ as $c(w)/(c(w) + \#(w))$ and $\lambda_C(\bar{q}(w)|w)$ as $\#(w)/(c(w) + \#(w))$ where $\#(w)$ is the total weight assigned to the novel events $\bar{q}(w)$ in the context $w$. Currently, we calculate $\#(w)$ as $\min(|q(w)|, |\bar{q}(w)|)$ so that highly variable contexts receive more flattening, but no novel symbol in $\bar{q}(w)$ receives more than unity weight. Next, $\lambda_C(\sigma|q(w), w)$ is estimated as $c(\sigma|w)/c(w)$ for the previously seen symbols $\sigma \in q(w)$ and $\lambda_C(\sigma|\bar{q}(w), w)$ is estimated uniformly as $1/|\bar{q}(w)|$ for the novel symbols $\sigma \in \bar{q}(w)$. Combining these estimates, we obtain our overall estimate (5).

$$\lambda_C(\sigma|w) = \begin{cases} \dfrac{c(\sigma|w)}{c(w) + \#(w)} & \text{if } \sigma \in q(w) \\ \dfrac{\#(w)}{|\bar{q}(w)|\,(c(w) + \#(w))} & \text{otherwise} \end{cases} \quad (5)$$

Unlike Moffat, our estimate (5) does not use escape probabilities or any other form of context blending. All novel events $\bar{q}(w)$ in the context $w$ are assigned uniform probability. This is suboptimal but simpler.

We note that our frequencies are incorrect when used in an extension model that contains contexts that are proper suffixes of each other. In such a situation, the shorter context is only used when the longer context was *not* used. Let $y$ and $xy$ be two distinct contexts in a model $\phi$. Then the context $y$ will never be used when the history is $\Sigma^* xy$. Therefore, our estimate of $\lambda(\cdot|y)$ should be conditioned on the fact that the longer context $xy$ did not occur. The interaction between candidate contexts can become quite complex, and we consider this problem in other work (Ristad and Thomas, 1995).

Parameter estimation is only a small part of the overall model estimation problem. Not only do we have to estimate the parameters for a model, we have to find the right parameters to use! To do this, we proceed in two steps. First, in section 2.3, we use the minimum description length (MDL) principle to quantify the total merit of a model with respect to a training corpus. Next, in section 2.4, we use our MDL codelengths to derive a practical model selection algorithm with which to find a good model in the vast class of all extension models.

## 2.3 Codelength Formulas

The goal of this section is to establish the proper tension between model complexity and data complexity, in the fundamental units of information. Although the MDL framework obliges us to propose particular encodings for the model and the data, our goal is not to actually encode the data or the model.

Given an extension model $\phi$ and a text corpus $T$, $|T| = t$, we define the total codelength $L(T, \phi|\Phi)$ relative to the model class $\Phi$ using a 2-part code.

$$L(T, \phi|\Phi) = L(\phi|\Phi) + L(T|\phi, \Phi)$$

Since conditioning on the model class $\Phi$ is always understood, we will henceforth suppress it in our notation.

Firstly, we will encode the text $T$ using the probability model $\phi$ and an arithmetic code, obtaining the following codelength.

$$L(T|\phi) = -\log \tilde{p}(T|\phi)$$

Next, we encode the model $\phi$ in three parts: the context dictionary as $L(D)$, the extensions as $L(E|D)$, and the conditional frequencies $c(\cdot|\cdot)$ as $L(c|D, E)$.

The dictionary $D$ of contexts forms a suffix tree containing $n_i$ vertices with branching factor $i$. The tree contains $n = \sum_{i=1}^{m} n_i$ internal vertices and $n_0$ leaf vertices. There are $(n_0 + n_1 + \ldots + n_m - 1)!/n_0!n_1!\ldots n_m!$ such trees (Knuth, 1986:587). Accordingly, this tree may be encoded with an enumerative code using $L(D)$ bits.

$$\begin{aligned} L(D) =\ & L_Z(n) + \log \binom{n + m - 1}{m - 1} \\ & + \log \frac{(n_0 + n_1 + \ldots + n_m - 1)!}{n_0!n_1!\ldots n_m!} \\ & + \sum_{i=1}^{m-1} n_i \log \binom{m}{i} + L_{Z \leq}(|\lfloor D \rfloor|, n) \\ & + \log \binom{n + |\lfloor D \rfloor| - 1}{|\lfloor D \rfloor| - 1} \end{aligned}$$

where $\lfloor D \rfloor$ is the set of all contexts in $D$ that are proper suffixes of another context in $D$. The first term encodes the number $n$ of internal vertices using the Elias code. The second term encodes the counts $\{n_1, n_2, \ldots, n_m\}$. Given the frequencies of these internal vertices, we may calculate the number $n_0$ of leaf vertices as $n_0 = 1 + n_2 + 2n_3 + 3n_4 + \ldots + (m-1)n_m$. The third term encodes the actual tree (without labels) using an enumerative code. The fourth term assigns labels (ie., symbols from $\Sigma$) to the edges in the tree. At this point the decoder knows all contexts which are not proper suffixes of other contexts, ie., $D - \lfloor D \rfloor$. The fourth term encodes the magnitude of $\lfloor D \rfloor$ as an integer bounded by the number $n$ of internal vertices in the suffix tree. The fifth term identifies the contexts $\lfloor D \rfloor$ as interior vertices in the tree that are proper suffixes of another context in $D$.

Now we encode the symbols available in each context. Let $m_i$ be the number of contexts that have exactly $i$ extensions, ie., $m_i \doteq |\{w : |E(w)| = i\}|$. Observe that $\sum_{i=1}^{m} m_i = |D|$.

$$L(E|D) = \log \binom{|D| + m - 1}{m - 1} + \log \binom{|D|}{\{m_i\}}$$
$$+ \sum_{i=1}^{m} m_i \log \binom{m}{i}$$

The first term represents the encoding of $\{m_i\}$ while the second term represents the encoding $|E(w)|$ for each $w$ in $D$. The third term represents the encoding of $E(w)$ as a subset of $\Sigma$ for each $w$ in $D$.

Finally, we encode the frequencies $c(\sigma|w)$ used to estimate the model parameters

$$L(c|D, E) = L_Z(c(\epsilon)) + \sum_{w \in D} \log \binom{c(w) + |\lceil w \rceil|}{c(w)}$$
$$+ \sum_{w \in D} \log \binom{c(w) + |E(w)|}{|E(w)|}$$

where $\lceil y \rceil$ consists of all contexts that have $y$ as their maximal proper suffix, ie., all contexts that $y$ immediately dominates, and $\lfloor y \rfloor$ is the maximal proper suffix of $y$ in $D$, ie., the unique context that immediately dominates $y$. The first term encodes $|T|$ with an Elias code and the second term recursively partitions $c(w)$ into $c(\lceil w \rceil)$ for every context $w$. The third term partitions the context frequency $c(w)$ into the available extensions $c(E(w)|w)$ and the "unallocated frequency" $c(\Sigma - E(w)|w) = c(w) - c(E(w)|w)$ in the context $w$.

### 2.4 Model Selection

The final component of our contribution is a model selection algorithm for the extension model class $\Phi$. Our algorithm repeatedly refines the accuracy of our model in increasingly long contexts. Adding a new parameter to the model will decrease the codelength of the data and increase the codelength of the model. Accordingly, we add a new parameter to the model only if doing so will decrease the total codelength of the data *and* the model.

The incremental cost and benefit of adding a single parameter to a given context cannot be accurately approximated in isolation from any other parameters that might be added to that context. Accordingly, the incremental cost of adding the *set* $\Sigma'$ of extensions to the context $w$ is defined as (6) while the incremental benefit is defined as (7).

$$\Delta L_\phi(w, \Sigma') \doteq L(\phi \cup (\{w\} \times \Sigma')) - L(\phi) \qquad (6)$$
$$\Delta L_T(w, \Sigma') \doteq L(T|\phi) - L(T|\phi \cup (\{w\} \times \Sigma')) \quad (7)$$

Keeping only significant terms that are monotonically nondecreasing, we approximate the incremental cost $\Delta L_\phi(w, \Sigma')$ as

$$\Delta L_\phi(w, \Sigma') \approx \log |D| + \log \binom{m}{|\Sigma'|}$$
$$+ \log c(\lfloor w \rfloor) + \log \binom{c(w) + |\Sigma'|}{|\Sigma'|}$$

The first term represents the incremental increase in the size of the context dictionary $D$. The second term represents the cost of encoding the candidate extensions $E(w) = \Sigma'$. The third term represents (an upper bound on) the cost of encoding $c(w)$. The fourth term represents the cost of encoding $c(\cdot|w)$ for $E(w)$. Only the second and fourth terms are signficant.

Let us now consider the incremental benefit of adding the extensions $\Sigma'$ to a given context $w$. The addition of a single parameter $\langle w, \sigma \rangle$ to the model $\phi$ will immediately change $\lambda(\sigma|w)$, by definition of the model class. Any change to $\lambda(\cdot|w)$ will also change the expansion factor $\delta(w)$ in that context, which may in turn change the conditional probabilities $\tilde{p}(\Sigma - E(w)|w, \phi)$ of symbols not available in that context. Thus the incremental benefit of adding the extensions $\Sigma'$ to the context $w$ may be calculated as

$$\Delta L_T(w, \Sigma') = c(\Sigma - \Sigma'|w) \log \frac{1 - \lambda(\Sigma'|w)}{1 - \tilde{p}(\Sigma'|w, \phi)}$$
$$+ \sum_{\sigma' \in \Sigma'} c(\sigma'|w) \log \frac{\lambda(\sigma'|w)}{\tilde{p}(\sigma'|w, \phi)}$$

The first term represents the incremental benefit (in bits) for evaluating $\Sigma - \Sigma'$ in the context $w$ using the more accurate expansion factor $\delta(w)$. The second term represents the incremental benefit (in bits) of using the direct estimate $\lambda(\sigma'|w)$ instead of the model probability $\tilde{p}(\sigma'|w, \phi)$ in the context $w$. Note that $\lambda(\sigma'|w)$ may be more or less than $\tilde{p}(\sigma'|w, \phi)$.

Now the incremental cost and benefit of adding a *single* extension $\langle w, \sigma \rangle$ to a model that already contains the extensions $\langle w, \Sigma' \rangle$ may be defined as follows.

$$\Delta L_\phi(w, \Sigma', \sigma) \doteq \Delta L_\phi(w, \Sigma' \cup \{\sigma\}) - \Delta L_\phi(w, \Sigma')$$

$$\Delta L_T(w, \Sigma', \sigma) \doteq \Delta L_T(w, \Sigma' \cup \{\sigma\}) - \Delta L_T(w, \Sigma')$$

Let us now use these incremental cost/benefit formulas to design a simple heuristic estimation algorithm for the extension model. The algorithm consists of two subroutines. Refine($D,E,n$) augments the model with all individually profitable extensions of contexts of length $n$. It rests on the assumption that adding a new context does not change the model's performance in the shorter contexts. Extend($w$) determines all profitable extensions of the candidate context $w$, if any exist. Since it is not feasible to evaluate the incremental profit of every subset of $\Sigma$, Extend($w$) uses a greedy heuristic that repeatedly augments the set of profitable extensions of $w$ by the single most profitable extension until it is not longer profitable to do so.

Refine($D,E,n$)
1.   $D_n := \{\}; E_n := \{\}$;
2.   $C_n := \{w : w \in C_{n-1}\Sigma \wedge c(w) > c_{\min}\}$;
3.   if $((n > n_{max}) \vee (|C_n| = 0))$ then return;
4.   for $w \in C_n$
5.       $S := \mathsf{Extend}(w)$;
6.       if $|S| > 0$ then $D_n := D_n \cup \{w\}; E_n(w) := S$;
7.   $D := D \cup D_n; E := E \cup E_n$;
8.   Refine($D,E,n+1$);

$C_n$ is the set of candidate contexts of length $n$, obtained from the training corpus. $D_n$ is the set of profitable contexts of length $n$, while $E_n$ is the set of profitable extensions of those contexts.

Extend($w$)
1.   $S := \{\}$;
2.   $\sigma := \mathsf{argmax}_{\sigma \in \Sigma} \{\Delta L(w, \{\sigma\})\}$
3.   while $(\Delta L(w, S, \sigma) > 0)$
4.       $S := S \cup \{\sigma\}$;
5.       $\sigma := \mathsf{argmax}_{\sigma \in \Sigma - S} \{\Delta L(w, S, \sigma)\}$
6.   return($S$);

The loop in lines 3-5 repeatedly finds the single most profitable symbol $\sigma$ with which to augment the set $S$ of profitable extensions. The incremental profit $\Delta L(\ldots)$ is the incremental benefit $\Delta L_T(\ldots)$ minus the incremental cost $\Delta L_\phi(\ldots)$.

Our breadth-first search considers shorter contexts before longer ones, and consequently the decision to add a profitable context $y$ may significantly decrease the benefit of a more profitable context $xy$, particularly when $c(xy) \approx c(y)$. For example, consider a source with two hidden states. In the first state, the source generates the alphabet $\Sigma = \{0, 1, 2\}$ uniformly. In the second state, the source generates the string "012" with certainty. With appropriate state transition probabilities, the source generates strings where $c(0) \approx c(1) \approx c(2)$, $c(2|1)/c(1) \gg c(2|\epsilon)/c(\epsilon)$, and $c(2|01)/c(01) > c(2|1)/c(1)$. In such a situation, the best context model includes the contexts "0" and "01" along with the empty context $\epsilon$. However, the divergence heuristic will first determine that the context "1" is profitable relative to the empty context, and add it to the model. Now the profitability of the better context "01" is reduced, and the divergence heuristic may therefore not include it in the model. This problem is best solved with a best first search. Our current implementation uses a breadth first search to limit the computational complexity of model selection.

Finally, we note that our parameter estimation techniques and model selection criteria are comparable in computational complexity to Rissanen's context models (1983, 1986). For that reason, extension models should be amendable to efficient online implementation.

## 3 Empirical Results

By means of the following experiments, we hope to demonstrate the utility of our context modeling techniques. All results are based on the Brown corpus, an eclectic collection of English prose drawn from 500 sources across 15 genres (Francis and Kucera, 1982). The irregular and nonstationary nature of this corpus poses an exacting test for statistical language models. We use the first 90% of each file in the corpus to estimate our models, and then use the remaining 10% of each file in the corpus to evaluate the models. Each file contains approximately 2000 words. Due to limited computational resources, we set $n_{\max} = 10$, $c_{\min} = 8$, and restrict our our alphabet size to 70 (ie., all printing ascii characters, ignoring case distinction).

Our results are summarized in the following table. Message entropy (in bits/symbol) is for the testing corpus only, as per traditional model validation methodology. The nonmonotonic extension model (NEM) outperforms all other models for all orders using vastly fewer parameters. Its performance all the more impressive when we consider that no context blending or escaping is performed, even for novel events.

We note that the test message entropy of the n-gram model class is minimized by the 5-gram at 2.38 bits/char. This result for the 5-gram is not honest because knowledge of the test set was used to select the optimal model order. Jelinek and Mercer (1980) have shown to interpolate n-grams of different order using mixing parameters that are conditioned on the history. Such interpolated Markov sources are considerably more powerful than traditional n-grams but contain even more parameters.

The best reported results on the Brown Corpus are 1.75 bits/char using a large interpolated trigram word model whose parameters are estimated using over 600,000,000 words of proprietary training data (Brown et.al., 1992). The use of proprietary training data means that these results are not independently repeatable. In contrast, our results were obtained using only 900,000 words of generally available training data and may be independently verified by any-one with the inclination to do so. The amount of training data is known to be a significant factor in model performance. Given a sufficiently rich dictionary of words and a sufficiently large training corpus, a model of word sequences is likely to outperform an otherwise equivalent model of character sequences. For these three reasons – repeatability, training corpus size, and the advantage of word models over character models – the results reported by Brown et.al (1992) are not directly comparable to those reported here.

| Model | Parameters | Entropy |
|---|---|---|
| NEM | 89,325 | 1.97 |
| NCM | 687,276 | 2.19 |
| $MCM_1$ | 88,945,904 | 2.43 |
| $MCM_2$ | 88,945,904 | 3.12 |
| n-gram | 506,352,021,176,052 | 3.74 |

Table 1: Results for the nonmonotonic extension model (NEM), the nonmonotonic context model (NCM), Rissanen's (1983,1986) monotonic context models ($MCM_1$, $MCM_2$) and the n-gram model. All models are order 7. The rightmost column contains test message entropy in bits/symbol. NEM outperforms all other model classes for all orders using significantly fewer parameters. It is possible to reduce the test message entropy of the NEM and NCM to 1.91 and 1.99, respectively, by quadrupling the number of model parameters.

Section 3.1 compares the statistical efficiency of the various context model classes. Next, section 3.2 anecodatally examines the complex interactions among the parameters of an extension model.

### 3.1 Model Class Comparison

Given the tremendous risk of overfitting, the most important property of a model class is arguably its statistical efficiency. Informally, statistical efficiency measures the effectiveness of individual parameters in a given model class. A high efficiency indicates that our model class provides a good description of the data. Conversely, a low efficiency indicates that the model class does not adequately describe the observed data.

In this section, we compare the statistical efficiency of three model classes: context models, extension models, and fixed-length Markov processes (ie., n-grams). Our model class comparison is based on three criteria of statistical efficiency: total codelength, bits/parameter on the test message, and bits/order on the test message. The context and extension models are all of order 9, and were estimated using the true incremental benefit and a range of fixed incremental costs (between 5 and 25 bits/extension for the extension model and between 25 and 150 bits/context for the context model).

According to the first criteria of statistical efficiency, the best model is the one that achieves the smallest total codelength $L(T, \phi)$ of the training corpus $T$ and model $\phi$ using the fewest parameters. This criteria measures the statistical efficiency of a model class according to the MDL framework, where we would like each parameter to be as cheap as possible and do as much work as possible. Figure 1 graphs the number of model parameters required to achieve a given total codelength for the training corpus and model. The extension model class is the overwhelming winner.

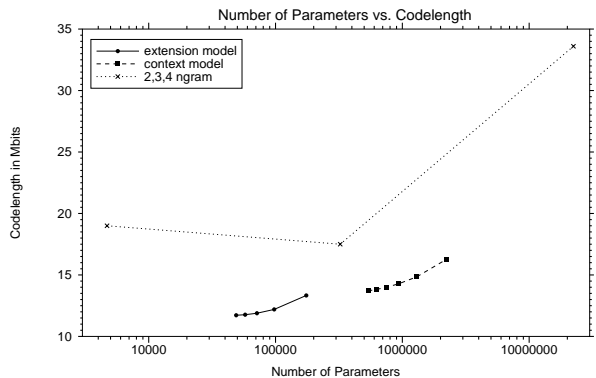

Figure 1: The relationship between the number of model parameters and the total codelength $L(T, \phi)$ of the training corpus $T$ and the model $\phi$. By this criteria of statistical efficiency, the extension models completely dominate context models and n-grams.

According to the second criteria of statistical efficiency, the best model is the one that achieves the lowest test message entropy using the fewest parameters. This criteria measures the statistical efficiency of a model class according to traditional model validation methodology, tempered by a healthy concern for overfitting. Figure 2 graphs the number of model parameters required to achieve a given test message entropy for each of the three model classes. Again, the extension model class is the clear winner. (This is particularly striking when the number of parameters is plotted on a linear scale.) For example, one of our extension models saves 0.98 bits/char over the trigram while using less than 1/3 as many parameters. Given the logarithmic nature of codelength and the scarcity of training data, this is a significant improvement.

According to the third criteria of statistical efficiency, the best model is one that achieves the lowest test message entropy for a given model order. This criteria is widely used in the language modeling community, in part because model order is typi-

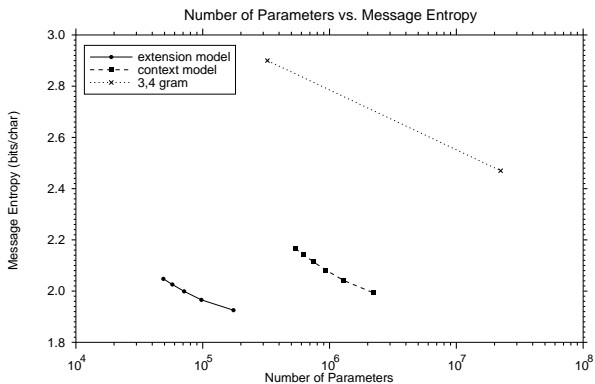

Figure 2: The relationship between the number of model parameters and test message entropy. The most striking fact about this graph is the tremendous efficiency of the extension model.

cally — although not necessarily — related to both the number of model parameters and the amount of computation required to estimate the model. Figure 3 compares model order to test message entropy for each of the three model classes. As the order of the models increases from 0 (ie., unigram) to 10, we naturally expect the test message entropy to approach a lower bound, which is itself bounded below by the true source entropy. By this criteria, the extension model class is better than the context model class, and both are significantly better than the n-gram.

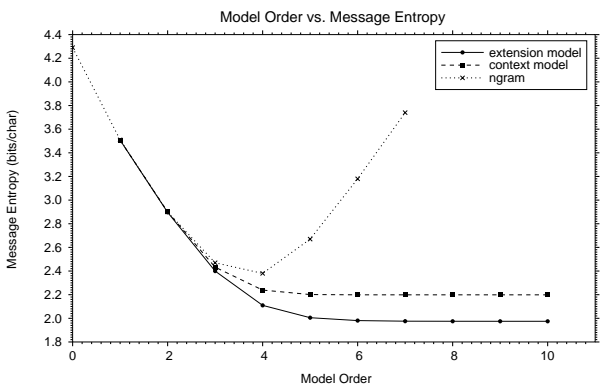

Figure 3: The relationship between model order and test message entropy. The extension model class is the clear winner by this criteria as well.

### 3.2 Anecdotes

It is also worthwhile to interpret the parameters of the extension model estimated from the Brown Corpus, to better understand the interaction between our model class and our heuristic model selection algorithm. According to the divergence heuristic, the decision to add an extension $\langle w, \sigma \rangle$ is made relative to that context's maximal proper suffix $\lfloor w \rfloor$ in $D$ as well as any other extensions in the context $w$. An extension $\langle w, \sigma \rangle$ will be added only if the direct estimate of its conditional probability is significantly different from its conditional probability in its maximal proper suffix after scaling by the expansion factor in the context $w$, ie., if $\lambda(\sigma|w)$ is significantly different than $\delta(w)\tilde{p}(\sigma|\lfloor w \rfloor)$.

This is illusrated by the three contexts and six extensions shown immediately below, where $+E(w)$ includes all symbols in $E(w)$ that are more likely in $w$ than they were in $\lfloor w \rfloor$ and $-E(w)$ includes all symbols in $E(w)$ that are less likely in $w$ than they were in $\lfloor w \rfloor$.

| $w$ | $+E(w)$ | $-E(w)$ |
|---|---|---|
| "blish" | e,i,m | |
| "o␣establish" | ␣ | |
| "e␣establish" | m | e |

The substring *blish* is most often followed by the characters 'e', 'i', and 'm', corresponding to the relatively frequent word forms *publish{ed, er, ing}* and *establish{ed, ing, ment}*. Accordingly, the context "blish" has three positive extensions {e,i,m}, of which e has by far the greatest probability. The context "blish" is the maximal proper suffix of two other contexts in the model, "o␣establish" and "e␣establish".

The substring *o establish* occurs most frequently in the gerund *to establish*, which is nearly always followed by a space. Accordingly, the context "o␣establish" has a single positive extension "␣". The substring *o establish* is also found before the characters 'm', 'e', and 'i' in sequences such as *to establishments*, *{who, ratio, also} established*, and *{to, into, also} establishing*. Accordingly, the context "o␣establish" does not have any negative extensions.

In contrast, the substring *e establish* is overwhelmingly followed by the character 'm', rarely followed by 'e', and never followed by either 'i' or space. For this reason, the context "e␣establish" has a single positive extension {m} corresponding to the great frequency of the string *the establishment*. This context also has single negative extension {e}, corresponding to the fact that the character 'e' is still possible in the context "e␣establish" but considerably less likely than in that context's maximal proper suffix "blish".

Since 'i' is reasonably likely in the context "blish" but completely unlikely in the context "e␣establish", we may well wonder why the model does not include the negative extension 'i' in addition to 'e' or even instead of 'e'. This puzzle is explained by the expansion factor as follows. Since the substring $e\ establish$ is only followed by 'm' and 'e', the expansion factor $\delta(\text{"e}_\sqcup\text{establish"})$ is essentially zero after 'm' and 'e' are added to that context, and therefore $\tilde{p}(\Sigma - \{m, e\}|\ \text{"e}_\sqcup\text{establish"})$ is also essentially zero. Thus, 'i' and space are both assigned nearly zero probability in the context $\text{"e}_\sqcup\text{establish"}$, simply because 'm' and 'e' get nearly all the probability in that context.

## 4 Conclusion

In ongoing work, we are investigating extension mixture models as well as improved model selection algorithms. An extension mixture model is an extension model whose $\lambda(\sigma|w)$ parameters are estimated by linearly interpolating the empirical probability estimates for all extensions that dominate $w$ with respect to $\sigma$, ie., all extensions whose symbol is $\sigma$ and whose context is a suffix of $w$. Extension mixing allows us to remove the uniform flattening of zero frequency symbols in our parameter estimates (5). Preliminary results are promising. The idea of context mixing is due to Jelinek and Mercer (1980).

Our results highlight the fundamental tension between model complexity and data complexity. If the model complexity does not match the data complexity, then both the total codelength of the past observations *and* the predictive error increase. In other words, simply increasing the number of parameters in the model does not necessarily increase predictive power of the model. Therefore, it is necessary to have a a fine-grained model along with a heuristic model selection algorithm to guide the expansion of the model in a principled manner.

**Acknowledgements.** Thanks to Andrew Appel, Carl de Marken, and Dafna Scheinvald for their critique. The paper has benefited from discussions with the participants of DCC95. Both authors are partially supported by Young Investigator Award IRI-0258517 to the first author from the National Science Foundation. The second author is additionally supported by a tuition award from the Princeton University Research Board. The research was partially supported by NSF SGER IRI-9217208.